\documentclass[runningheads,a4paper]{llncs}
\usepackage[utf8]{inputenc}
\usepackage{times}
\usepackage{amssymb}
\usepackage{graphicx}
\usepackage{multirow}
\usepackage{color}
\usepackage{booktabs}
\usepackage{url}
\usepackage[section]{placeins}
\usepackage{array}

\newcommand{\keywords}[1]{\par\addvspace\baselineskip
	\noindent\keywordname\enspace\ignorespaces#1}

\urldef{\mailsa}\path|leandrobs@usp.br, {nathansh, sandra}@icmc.usp.br,magali@uol.com.br|
\urldef{\mailsaa}\path|{leandrobs}@usp.br|
\urldef{\mailsb}\path|arnaldoc@utfpr.edu.br|
\urldef{\mailsc}\path|g.h.paetzold@sheffield.ac.uk|

\begin{document}
	
	\titlerunning{Automatically Inferring Psycholinguistic Properties of Brazilian Portuguese}
	
	\title{A Lightweight Regression Method to Infer Psycholinguistic Properties for Brazilian Portuguese}
	
	\author{Leandro Borges dos Santos\inst{1} \and Magali Sanches Duran\inst{1} \and Nathan Siegle Hartmann\inst{1} \and Arnaldo Candido Jr.\inst{2} \and Gustavo Henrique  Paetzold\inst{3} \and Sandra Maria Aluisio\inst{1}}

	\institute{University of S\~{a}o Paulo,  Institute of Mathematics and Computer Sciences\\
		\mailsa\\
		\and
		Federal Technological University of Paran\'{a}  (UTFPR), Medianeira \\
		\mailsb\\
		\and
		University of Sheffield, Department of Computer Science \\
		\mailsc\\
	}

	\index{dos Santos, Leandro}
	\index{Duran, Magali}
	\index{Hartmann, Nathan}
	\index{Candido Jr., Arnaldo}
	\index{Paetzold, Gustavo}
	\index{Aluisio, Sandra}
	
	\toctitle{} \tocauthor{}
	
	\maketitle
%
%
%
%
\begin{abstract}
Psycholinguistic properties of words have been used in various approaches to Natural Language Processing tasks, such as text simplification and readability assessment. Most of these properties are subjective, involving costly and time-consuming surveys to be gathered. Recent approaches use the limited datasets of psycholinguistic properties to extend them automatically to large lexicons. However, some of the resources used by such approaches are not available to most languages. This study presents a method to infer psycholinguistic properties for Brazilian Portuguese (BP) using regressors built with a light set of features usually available for less resourced languages: word length, frequency lists, lexical databases composed of school dictionaries and word embedding models. The correlations between the properties inferred are close to those obtained by related works. The resulting resource contains 26,874 words in BP annotated with concreteness, age of acquisition, imageability and subjective frequency.

\end{abstract}

\normalsize
\section{Introduction}

Besides frequency, form, and meaning, words also have several other less known words properties, such as imageability, concreteness, familiarity, subjective frequency, and age of acquisition (AoA), which are subjective psycholinguistic properties, as they depend on the personal experiences that individuals had using those words. 
According to \cite{soares:2016:mwp}, word imageability is the ease and speed with which a word evokes a mental image; concreteness is the degree to which words refer to objects, people, places, or things that can be experienced by the senses; experiential familiarity is the degree to which individuals know and use words in their everyday life; subjective frequency is the estimation of the number of times a word is encountered by individuals in its written or spoken form, and AoA is the estimation of the age at which a word was learned.
Psycholinguistic properties have been used in various approaches, such as for Lexical Simplification \cite{paetzold:2016:inferring}, for Text Simplification at the sentence level, with the aim of reducing the difficulty of informative text for language learners \cite{DBLP:journals/corr/VajjalaM16}, to predict the reading times (RTs) of each word in a sentence to assess sentence complexity \cite{Singh2016QuantifyingSC} and also to create robust text level readability models \cite{Vajjala2014}, which is also one of the purposes of this paper.

Because of its inherent costs, the measurement of subjective psycholinguistic properties is usually used in the creation of datasets of limited size \cite{cameirao:2010:aoa,janczura:2007:concretude,marques:2007:aoa,soares:2016:mwp}. 
For the English language, the most well known database of this kind is the MRC Psycholinguistic Database \footnote{\url{websites.psychology.uwa.edu.au/school/MRCDatabase/mrc2.html}}, which contains 27 subjective psycholinguistic properties for 150,837 words. 
For BP for example, there is a psycholinguistic database \footnote{ \url{www.lexicodoportugues.com}} containing 21 columns of information for 215,175 words, but no subjective psycholinguistic properties. 

In this work we aim to overcome this gap by automatically inferring the psycholinguistic properties of imageability, concreteness, AoA and subjective  frequency (similar to familiarity) for a large database of 26,874 BP words using a resource-light regression approach.  As for the automatic inference, this work is strongly based on the results of \cite{paetzold:2016:inferring}  which proposed an automatic bootstrapping method for regression to populate the MRC Database. 
We explore here 3 research questions: 
(1) is it possible to achieve high Pearson and Spearman correlations values and low MSE values with a regression method using only word embedding features to infer the psycholinguistic properties for BP? 
(2) which size a database with psycholinguistic properties should have to be used in regression models? Does merging databases from different sources yield better correlation and lower MSE scores? 
(3) can the inferred values help in creating features that result in more reliable readability prediction models of BP texts for early school years (from 3rd to 6th grades)? 
Moreover, we assessed interrater reliability (Cronbach’s alpha) between ratings generated by our method and 
the imageability and concreteness 
produced for 237 nouns 
by \cite{marques:2005:normas}. Besides that, we analyzed the relations between the inferred ratings and other psycholinguistic variables.

\section{Related Works}

To the best of our knowledge there are only two studies that propose regression methods to automatically estimate missing psycholinguistic properties in the MRC Database \cite{conf/flairs/FengCCM11,paetzold:2016:inferring}.
In order to solve limitations resulting from using word databases with human ratings, \cite{conf/flairs/FengCCM11} proposes a computational model to predict word concreteness, by using linear regression with word attributes from WordNet \cite{fellbaum1998}, Latent Semantic Analysis (LSA) and the CELEX Database\footnote{\url{celex.mpi.nl/}} and use these attributes to simulate human ratings in the MRC database. Word concreteness is among the most important indices provided by Coh-Metrix, as comprehension is facilitated by virtue of more concrete words. The lexical features used were 19 lexical types from WordNet, 17 LSA dimensions, hypernymy information from WordNet, word frequencies from the CELEX Database, and word length (i.e., number of letters), totalling 39 attributes. The Pearson correlation between the estimated concreteness score and the concreteness score in the test set was 0.82. 

\cite{paetzold:2016:inferring} automatically estimate missing psycholinguistic properties in the MRC Database through a bootstrapping algorithm for regression. Their method exploits word embedding models and 15 lexical features, including the number of senses, synonyms, hypernyms and hyponyms for word in WordNet and also minimum, maximum and average distance between the word’s senses in WordNet and the thesaurus’ root sense. The Pearson correlation  between the estimated score and the inferred score for familiarity was 0.846; 0.862 for AoA; 0.823 for imagenery and 0.869 for concretness, which is better than the results of \cite{conf/flairs/FengCCM11}.

\section{A Lightweight Regression Method to Infer Psycholinguistic Properties of Words\label{sec:lightweight:regression}}

The fact that the methods developed by \cite{conf/flairs/FengCCM11} and \cite{paetzold:2016:inferring} are based on a large, scarce lexical resources as WordNet, led us to raise the question ``Could we have a similar performance with a simpler set of features which are easily obtainable for most languages?''. Therefore we decided to build our regressors using only word length, frequency lists, lexical databases composed of school dictionaries and word embeddings models.
One critical difference between the strategy of \cite{paetzold:2016:inferring} and ours is that they concatenate all features to train a regressor, while we take a different approach. Although simply combining all features is straightforward, it can lead to noise insertion, given that the features used greatly contrast among them (e.g. word embeddings and word length). Instead, we adopted a more elegant solution, called Multi-View Learning~\cite{xu:2013:survey:mv}. In a Multi-View Learning, multiple regressors/classifiers are trained over different feature spaces and then combined to produce a single result. Here, the fusion stage is made by averaging the values predicted by the regressors \cite{xu:2013:survey:mv}.

\subsection{Adaptation of Databases with Psychological Norms for Portuguese Words\label{sec:adaptation:databases}}

We present in Table \ref{table:norms} surveys involving the subjective psycholinguistic properties of words focused in this study (concreteness, age of acquisition, imageability and subjective frequency), both for European Portuguese (EP) and BP.
\begin{table}[ht]
\centering
\scalebox{.75}{
\begin{tabular}{@{}c|c|c|c|c|c@{}}
\hline
Study                                                               & \begin{tabular}[c]{@{}l@{}}Number\\ of Participants\end{tabular} & \begin{tabular}[c]{@{}l@{}}Number\\ of Words\end{tabular} & Property                                                                                   & Variant & Scale \\ \hline
\begin{tabular}[c]{@{}l@{}}\cite{soares:2016:mwp}\end{tabular}        & 2,357                                                             & 3,789                                                      & \begin{tabular}[c]{@{}l@{}}concreteness, imageability, subjective frequency\end{tabular} & EP      & 1-7   \\ \hline
\begin{tabular}[c]{@{}l@{}}\cite{cameirao:2010:aoa}\end{tabular} & 685                                                              & 1,748                                                      & AoA                                                                                        & EP      & 1-9   \\ \hline
\begin{tabular}[c]{@{}l@{}}\cite{janczura:2007:concretude}\end{tabular}      & 719                                                              & 909                                                       & concreteness                                                                               & BP      & 1-7   \\ \hline
\begin{tabular}[c]{@{}l@{}}\cite{marques:2007:aoa}\end{tabular}       & 110                                                              & 834                                                       & AoA                                                                                        & EP      & 1-7   \\ \hline
\begin{tabular}[c]{@{}l@{}}\cite{marques:2005:normas}\end{tabular}                                                        & 103                                                              & 249                                                       & \begin{tabular}[c]{@{}l@{}}imageability,  concreteness\end{tabular}                                                              & EP      & 1-7   \\ \hline
\end{tabular}}
\caption{Norms for Portuguese on the focused psycholinguistic properties.}
\label{table:norms}
\end{table}

If, on the one hand, manually produced resources fulfill the needs for which they were collected, on the other hand, they are very limited for Natural Language Processing purposes, given their limited size. 
There is place, however, to automatically infer subjective psycholinguistic properties for several words, using the existing ones. To achieve this goal, however, we need, first of all, to rely on a set of words with values for each aimed property. 

In BP, we only have 909 words with concreteness values \cite{janczura:2007:concretude}. Therefore, we decided to incorporate EP resources to our set, as well as to combine different resources containing values for the same property. In order to turn EP resources usable for our study in BP, we executed adjustments in the word lists. Most of them were in orthography, as for example: ac\c{c}\~{a}o/ac\~{a}o (action), adop\c{c}\~{a}o/ado\c{c}\~{a}o (adoption), amnistia/anistia (amnesty). Other adjustments pertained to concepts that the two variants of Portuguese lexicalize in different ways, such as: ficheiro/arquivo (file), assass\'{\i}nio
/assassinato (murder), apuramento/apura\c{c}\~{a}o (calculation). Finally, some words have been discarded, as they lexicalize concepts related to fauna, flora, and culinary traits native to Portugal, such as: faneca (pout), faia (beech) and rebu\c{c}ado (candy). 
In theory, there should not be a problem in concatenating two or more lists of words with the same psychological property. We did this for concreteness, merging the list of \cite{soares:2016:mwp}, once adapted fo BP, with the one of \cite{janczura:2007:concretude}, which was created for BP. As both lists rated concreteness using a Likert scale of 7 points, the values were comparable. However, in what concerns AoA, the two lists available, \cite{cameirao:2010:aoa} and \cite{marques:2007:aoa}, rated concreteness using respectively a Likert scale of 7 and 9 points. It is worth mentioning that both lists obtained AoA ratings through estimates produced by adults (AoA could be, alternatively, gathered using the proficiency of children of different ages in object naming tasks). Therefore, to turn them comparable, we had to convert the scale of 9 points into a scale of 7 points. 
After concluding the lexical adjustments, converting the scale of 9 points into 7 points for AoA lists, merging the lists and eliminating duplicated words, we obtained sizeable datasets for all word properties addressed in this study. Table \ref{regressor:results} shows the number of entries obtained for each property, between the parentheses.

\subsection{Features \label{sec:features}}

Our regressors use 10 features from several sources, grouped in: (i) lexical (1-8); (ii) Skip-Gram word embeddings (9) \cite{mikolov:2013:efficient}; and (iii) GloVe word embeddings (10) \cite{pennington:2014:glove}:

\begin{enumerate}
	\item Log of Frequency in SUBTLEX-pt-BR ~\cite{tang:2012:million:word}, which is a database of BP word frequencies based on more than 50 million words from film and television subtitles;
    \item Log of Contextual diversity in SUBTLEX-pt-BR, which is the number of subtitles that contain the word;
	\item Log of Frequency in SubIMDb-PT~\cite{paetzold-specia:2016:COLING2}: this corpus was extracted from subtitles of family, comedy and children movies and series;
    \item Log of Frequency in the Written Language part of Corpus Brasileiro, a corpus with about 1 billion words of Contemporary BP; 
    \item Log of Frequency in the Spoken Language part of Corpus Brasileiro;
	\item Log of Frequency in a corpus of 1.4 billion tokens of Mixed Text Genres in BP;
    \item Word Length;
    \item Lexical databases from 6 school dictionaries for specific grade-levels;
  \item Word's raw embedding values of word embeddings models created using the Skip-Gram algorithm \cite{mikolov:2013:efficient}, with word vector sizes of 300, 600 and 1,000;
\item Word's raw embedding values of word embeddings models created using the GloVe algorithm \cite{pennington:2014:glove}, with word vector sizes of 300, 600 and 1,000;
\end{enumerate}

Reading time studies provide evidence that more processing time is allocated to rare words than high-frequency words. Besides that, the logarithm of word frequency was used here because reading times are linearly related to the logarithm of word frequency, not to raw word frequencies \cite{graesser2011}.
We trained our embedding models using Skip-Gram word2vec 
and GloVe over a corpus of 1.4 billion tokens, and 3.827.725 types composed by mixed text genres, including subtitles to cover spoken language besides written texts.

\subsection{\label{sec:regression:multi:view}Using Regression in a Multi-View Learning Approach}

We used  a linear least squares regressor with L2 regularization, which is also known  as Ridge Regression or Tikhonov regularization \cite{hoerl:1970:ridge}. We choose this regression method due to the promising results reported by~\cite{paetzold:2016:inferring}. We trained three regressors in different feature spaces: lexical features, Skip-Gram embeddings, and GloVe embeddings. 

\section{Evaluation}

We experimented with several dimensions of word embeddings, but for space reasons, here we include only the best results: Skip-Gram and GloVe embeddings with 300 word vector dimensions.
We used 20x5-fold cross-validation in order to perform our experiments. As evaluation metrics, we used Mean Square Error (MSE), Spearman's ($\rho$), and Pearson's ($r$) correlation. For the MSE metric, a repeated measures ANOVA with Dunnet post-test was used to compare the best regressors with the others to significance level of 0.05.
Table \ref{regressor:results} shows the evaluation results of our method. For subjective frequency, the best result was given by the combination of Lexical, Skip-gram and GloVe embeddings. For AoA, the best result was given by the combination of Lexical and GloVe embeddings. For AoA, the three better lexical features responsible for such results were grade-level lexical databases, the log frequency in Sub-IMDb-PT and word length. The regressors in bold presented statistically significant differences when compared with the others. However, for AoA property, the Lexical + GloVe regressor did not present difference statistically significant with Lexical + Skip-gram + GloVe regressor.

\begin{table}[!ht]
	\centering
    \scalebox{.75}{
	\begin{tabular}{@{}c|ccc|ccc|ccc|ccc@{}}
		\toprule
		\multirow{3}*{Regressors} & \multicolumn{3}{c|}{Concreteness} & \multicolumn{3}{c|}{Subjective Frequency} & \multicolumn{3}{c|}{Imageability} & \multicolumn{3}{c}{AoA Merging} \\
		
		& \multicolumn{3}{c|}{(4088)} & \multicolumn{3}{c|}{(3735)} & \multicolumn{3}{c|}{(3735)} & \multicolumn{3}{c}{(2368)}\\ \cmidrule(l){2-13}
		& MSE & $r$ & $\rho$ & MSE & $r$ & $\rho$ & MSE & $r$ & $\rho$ & MSE & $r$ & $\rho$ \\ \midrule
		Lexical & 1.24 & 0.54 & 0.56 & 0.55 & 0.72 & 0.73 & 0.74 & 0.58 & 0.59 & 0.67     & 0.73        & 0.73        \\
		Skip-gram & 0.52 & 0.84 & 0.84 & 0.58 & 0.70 & 0.71 & 0.46 & 0.77 & 0.77 & 0.81 & 0.66 & 0.66 \\
		GloVe & 0.62 & 0.80 & 0.81 & 0.40 & 0.81 & 0.81 & 0.49     & 0.75 & 0.75 & 0.63 & 0.75 & 0.75 \\ \midrule
		Lexical + Skip-gram & 0.64 & 0.82 & 0.82 & 0.44 & 0.79 & 0.79 & 0.47 & 0.77 & 0.78 & 0.59 & 0.77 & 0.77 \\
		Lexical + GloVe & 0.70 & 0.80 & 0.80 & 0.39 & 0.81 & 0.81 & 0.50 & 0.75 & 0.76 & \textbf{0.54} & \textbf{0.79} & \textbf{0.79} \\
		Skip-gram + GloVe & \textbf{0.49} & \textbf{0.85} &	\textbf{0.85} & 0.41 & 0.80 &	0.80 & \textbf{0.42} & \textbf{0.79} &	\textbf{0.79} & 0.62 & 0.75	& 0.75 \\
		Lexical + Skip-gram + GloVe & 0.55 & 0.85 & 0.84 & \textbf{0.38} & \textbf{0.82} & \textbf{0.82} & 0.43 & 0.79 & 0.78 & 0.54 & 0.79 & 0.79 \\ \bottomrule
	\end{tabular}}
	\caption{MSE and Pearson and Spearman correlation scores of the regression models. \label{regressor:results}}
\end{table}

Portuguese databases with AoA properties are small in size, therefore we evaluated three different databases for this property. The first has 765 words \cite{marques:2007:aoa}, the second has 1717 words \cite{cameirao:2010:aoa}, and the third is composed by a merging of the first and the second converted into a 7-scale; the merging resulted in a database with 2368 different words, which is still small compared to the other 3 properties evaluated here. The resulting correlations and MSE for AoA (see Table \ref{aoa:improvement}) show that merged datasets yield better results. There was a drop of 0.26 in MSE scores and an increase of 0.07 and 0.08 of Pearson and Spearman values.

\begin{table}[!ht]
\centering
\scalebox{.75}{
\begin{tabular}{@{}c|ccc|ccc|ccc@{}}
\toprule
\multirow{2}{*}{Regressors} & \multicolumn{3}{c|}{AoA (765)} & \multicolumn{3}{c|}{AoA (1717)} & \multicolumn{3}{c}{AoA Merge (2368)} \\ \cmidrule(l){2-10}
& MSE & $r$ & $\rho$ & MSE & $r$ & $\rho$ & MSE & $r$ & $\rho$ \\ \midrule
Lexical & 0.91 & 0.67 & 0.66 & 1.04 & 0.76 & 0.75 & 0.67 & 0.73 & 0.72\\
Skip-gram & 1.30 & 0.56 & 0.58 & 1.36 & 0.68 & 0.65 & 0.81 & 0.66 & 0.66\\
GloVe & 1.18 & 0.62 & 0.63 & 0.93 & 0.79 & 0.75 & 0.63 & 0.75 & 0.75\\
Lexical + GloVe & \textbf{0.80} & \textbf{0.72} & \textbf{0.71} & \textbf{0.79} &\textbf{0.83} & \textbf{0.80} & \textbf{0.54} & \textbf{0.79} & \textbf{0.79} \\ \bottomrule
\end{tabular}}
\caption{MSE, Pearson, and Spearman correlations of the regression models.\label{aoa:improvement}}
\end{table}

We also compared the four properties' interdependency among themselves by using Pearson correlation. Table \ref{tab:interdependency} presents the results obtained for our comparisons, as well as the results obtained on similar comparisons from related contributions. In Table \ref{tab:interdependency}, dashes represent evaluations which were not performed on a given study.
Our results are close to those reported in the literature, except for the correlation between age of aquisition and concreteness, which is stronger in other studies. This may be related to the fact that a full dicitionary have a larger proportion of rare words than observed in the lists of these studies.
\begin{table}[ht]
\centering
\scalebox{.75}{
\begin{tabular}{@{}c|c|c|c|c|c|cc@{}}
\toprule
  Properties compared & 
  OURS & 
  \cite{marques:2007:aoa} & 
  \cite{cameirao:2010:aoa} & 
  \cite{soares:2016:mwp} &
  \cite{marques:2007:aoa} vs  \cite{soares:2016:mwp}  &
  \cite{cameirao:2010:aoa} vs \cite{soares:2016:mwp}   \\ \hline

  AoA vs Concreteness          & -0.37 & -0.52 & -0.61  & -     & -0.49  & -0.54 \\ \hline
  AoA vs Imageability          & -0.73 & -0.69 & -0.66  & -     & -0.66  & -0.62 \\ \hline
  AoA vs Sub. Freq.            & -0.52 &    - &    - & -     & -0.65 & -0.60 \\ \hline
  Imageability vs Sub. Freq.   & 0.11  &    - &    - & 0.04  & -0.10  & -  \\ \hline
  Concreteness vs Sub. Freq.   & -0.05  &    - &    - & -0.09 & -     & - \\ \hline
  Imageability vs Concreteness & 0.92 &    - &    - & 0.88  & 0.82     & - \\ \hline
\end{tabular}}
\caption{Pearson correlations among properties.}
\label{tab:interdependency}
\end{table}
In order to validate the reliability of our automatically inferred psycholinguistic properties, we conducted internal consistency analyses. 
We calculated alpha scores between our automatically produced imageability and concreteness properties and from those present in the psycholinguistic dataset of \cite{marques:2005:normas}. In total, 237 words were considered. The alpha scores for imageability and concreteness are 0.921 and 0.820, which are similar to the values achieved by \cite{soares:2016:mwp}, and suggest that our features do, in fact, accurately capture the psycholinguistic properties being targeted.

We built a database of plain words which was populated with the inferred values for the four psycholinguistic properties focused in this study.
For this, we took profit of Minidicion{\'a}rio Caldas Aulete’s entries \cite{minidicionariocontemporaneo2010} and their respective first grammatical category. In the sequence, we selected only nouns, verbs, adjectives and adverbs. Finally, we eliminated all loanwords (foreign words from different origins used in BP). Then we searched the frequency of each word in the large corpus of 1.4 billion words we have used to train our word embeddings models, and after manual analysis, we decided to disregard words with less than 8 occurrences, as they are very uncommon. The final lexicon is available\footnote{\url{https://www.dropbox.com/s/evieeic673gomvn/PB.zip?dl=0}.} and contains 26,874 words, being 15,204 nouns, 4,305 verbs,  7,293 adjectives and 72 adverbs with the information of the four inferred psycholinguistic properties using the better results with less features (shown in bold in Table \ref{regressor:results}).

\section{Evaluating psycholinguistic features in readability prediction}

In order to evaluate the use of psycholinguistic properties to predict the readability level of BP informative texts from newspapers and magazines for early school years, we trained a classifier using a corpus of 1,413 texts which were classified as ease to read for 3rd to 6th graders. These texts were annotated by 2 linguists (0,914 weighted kappa) and the corpus distribution by grade levels is 183 texts for 3rd grade, 361 texts for 4th grade, 537 texts for 5th grade and 332 texts for 6th-grade year. We are still in the annotation process to have a dataset similar in size to the work by \cite{Vajjala2014}. 
We compared the use of our four psycholinguistic features and six traditional readability formulas: Flesch Reading-Ease adapted to BP Brun\'{e}t, Honor\'{e}, Dale-Chall, Gunning Fox and Moving Average Type-Token Ratio (MATTR). 
We used the mean and the average of our psycholinguistic properties as features to train SVM classifiers with RBF kernel for each psycholinguistic information and a single-view classifier for all four properties. All results presented here were obtained by a 10-fold cross-validation process.
Table \ref{tab:features_individuais_readability} shows that subjective frequency provided better results than the other psycholinguistic features in classifying grade levels and achieved the 3rd best result for individual features. The single-view classifier of psycholinguistic features overcame all traditional formulas but MATTR and Brun\'{e}t Index for grade-level classification in F1-measure.
Both MATTR and Brun\'{e}t Index measure the lexical diversity of a text and are independent of text length. Their high performance in this evaluation suggests that lexical diversity is a strong proxy to distinguish grade levels in primary school years.

\newcolumntype{C}[1]{>{\centering\let\newline\\\arraybackslash\hspace{0pt}}m{#1}}

\begin{table}
\center
\footnotesize
\scalebox{.75}{
\begin{tabular}
{c|c|c|c|c|c|C{.8cm}|C{1.2cm}|C{1.5cm}|C{1.4cm}|c|c}
\toprule
	Features & Flesch & Honor\'{e} & Concreteness & Familiarity & AoA & Dale-Chall & Gunning Fox & Subjective Frequency & Psycholinguistics & MATTR & Brun\'{e}t \\
    \midrule
    F1 & 0.26 & 0.29 & 0.27 & 0.23 & 0.25 & 0.36 & 0.37 & 0.32 & 0.45 & 0.48 & \textbf{0.54}\\
    \bottomrule
\end{tabular}}
\caption{Evaluating Psycholinguistic and Classic readability formulas for readability prediction.}
    \label{tab:features_individuais_readability}
\end{table}

\section{Conclusion and Future Work}

In this work, we set our aims at finding a light set of features available for most languages to build regressors that infer psycholinguistic properties for BP words. We have made publicly available a large database of 26,874 BP words annotated with psycholinguistic properties. 
With respect to our research questions (1) and (2), we have shown we can infer psycholinguistic properties for BP using word embeddings as features. Nonetheless, our regressors need a reasonably sized number of training instances (at least, more than two thousand examples), as well as complementary lexical resources to yield top performance for AoA and subjective frequency.
As for the research question (3), the results show that psycholinguistic properties have potential to help readability prediction. This results ratify the claims of \cite{Singh2016QuantifyingSC}, which state that  (i) words with higher concreteness are easier to imagine, comprehend, and memorize and therefore help readability of texts, and (ii) age of acquisition has been shown to be helpful in predicting reading difficulty.
As future work, we propose to extend evaluation to other tasks, to use new modelling of our psycholinguistic features (besides the average and standard deviation of the inferred values) and to use a more robust approach to perform the fusion of regressors, e.g. stacking regression.

\bibliographystyle{splncs03}
\bibliography{paper}

\begin{thebibliography}{10}
\providecommand{\url}[1]{\texttt{#1}}
\providecommand{\urlprefix}{URL }

\bibitem{graesser2011}
Arthur C.~Graesser, D.S.M.: Computational analyses of multilevel discourse
  comprehension. Topics in Cognitive Science  3(2),  371--98 (2011)

\bibitem{cameirao:2010:aoa}
Cameirao, M.L., Vicente, S.G.: Age-of-acquisition norms for a set of 1,749
  portuguese words. Behavior research methods  42(2),  474--480 (2010)

\bibitem{fellbaum1998}
Fellbaum, C.: {Wordnet: An Electronic Lexical Database}. MIT Press (1998)

\bibitem{conf/flairs/FengCCM11}
Feng, S., Cai, Z., Crossley, S.A., McNamara, D.S.: Simulating human ratings on
  word concreteness. In: FLAIRS Conference. AAAI Press (2011)

\bibitem{minidicionariocontemporaneo2010}
Geiger, P.: Minidicion\'{a}rio contempor\^{a}neo da l\'{\i}ngua portuguesa
  (2011)

\bibitem{hoerl:1970:ridge}
Hoerl, A.E., Kennard, R.W.: Ridge regression: Biased estimation for
  nonorthogonal problems. Technometrics  12(1),  55--67 (1970)

\bibitem{janczura:2007:concretude}
Janczura, G., Castilho, G., Rocha, N., van Erven, T., Huang, T.: Normas de
  concretude para 909 palavras da l\'{\i}ngua portuguesa. {Psicologia: Teoria e
  Pesquisa} pp. 195--204 (2007)

\bibitem{marques:2007:aoa}
Marques, J.F., Fonseca, F.L., Morais, S., Pinto, I.A.: Estimated age of
  acquisition norms for 834 portuguese nouns and their relation with other
  psycholinguistic variables. Behavior Research Methods pp. 439--444 (2007)

\bibitem{marques:2005:normas}
Marques, J.F.: Normas de imag{\'e}tica e concreteza para substantivos comuns.
  Laborat{\'o}rio de Psicologia  3,  65--75 (2005)

\bibitem{mikolov:2013:efficient}
Mikolov, T., Chen, K., Corrado, G., Dean, J.: Efficient estimation of word
  representations in vector space. In: Proceedings of ICLR 2013 (2013)

\bibitem{paetzold-specia:2016:COLING2}
Paetzold, G., Specia, L.: Collecting and exploring everyday language for
  predicting psycholinguistic properties of words. In: Proceedings of COLING
  2016. pp. 1669--1679 (2016)

\bibitem{paetzold:2016:inferring}
Paetzold, G.H., Specia, L.: Inferring psycholinguistic properties of words. In:
  Proceedings of NAACL-HLT. pp. 435--440 (2016)

\bibitem{pennington:2014:glove}
Pennington, J., Socher, R., Manning, C.D.: Glove: Global vectors for word
  representation. In: EMNLP 2014. pp. 1532--1543 (2014)

\bibitem{Singh2016QuantifyingSC}
Singh, A.D., Mehta, P., Husain, S., Rajkumar, R.: Quantifying sentence
  complexity based on eye-tracking measures (2016)

\bibitem{soares:2016:mwp}
Soares, A.P., Costa, A.S., J, M., Comesana, M.H.M.: The minho word pool: Norms
  for imageability, concreteness, and subjective frequency for 3,800 portuguese
  words. Behavior Research Methods  (2016)

\bibitem{tang:2012:million:word}
Tang, K.: A 61 million word corpus of brazilian portuguese film subtitles as a
  resource for linguistic research. UCL Work Pap Linguist  24,  208--214 (2012)

\bibitem{Vajjala2014}
Vajjala, S., Meurers, D.: Readability assessment for text simplification from
  analysing documents to identifying sentential simplifications. Recent
  Advances in Automatic Readability Assessment and Text Simplification  165(2),
   194–--222 (2014)

\bibitem{DBLP:journals/corr/VajjalaM16}
Vajjala, S., Meurers, D.: Readability-based sentence ranking for evaluating
  text simplification. CoRR  abs/1603.06009 (2016)

\bibitem{xu:2013:survey:mv}
Xu, C., Tao, D., Xu, C.: A survey on multi-view learning. arXiv preprint:
  1304.5634  (2013)

\end{thebibliography}

\end{document}